\documentclass{article}

\PassOptionsToPackage{square,numbers}{natbib}
\usepackage{amsmath}
\usepackage{amsfonts}
\usepackage{amssymb}
\usepackage{stackengine}
\usepackage{float}
\usepackage{wrapfig}
\usepackage[preprint]{neurips_data_2024}
\usepackage{caption} 
\usepackage[dvipsnames]{xcolor}
\usepackage{lipsum}
\PassOptionsToPackage{square,numbers}{natbib}
\usepackage{amsmath}
\usepackage{amsfonts}
\usepackage{mdframed}

\usepackage{amssymb}
\usepackage{stackengine}
\usepackage{float}
\usepackage{wrapfig}
\usepackage{neurips_data_2024}
\usepackage{caption} 
\usepackage{graphicx}

\usepackage[utf8]{inputenc} 
\usepackage[T1]{fontenc}    
\usepackage[colorlinks,citecolor=gray]{hyperref}        
\usepackage{url}            
\usepackage{booktabs}       
\usepackage{amsfonts}       
\usepackage{nicefrac}       
\usepackage{microtype}      
\usepackage{xcolor}         
\usepackage{graphicx}
\newcommand{\sectioncolor}{violet}

\newcommand{\myparagraph}[1]{\vspace{-2pt}\paragraph{#1}}

\usepackage{graphicx}

\usepackage[utf8]{inputenc} 
\usepackage[T1]{fontenc}    
\usepackage[colorlinks,citecolor=gray]{hyperref}        
\usepackage{url}            
\usepackage{booktabs}       
\usepackage{amsfonts}       
\usepackage{nicefrac}       
\usepackage{microtype}      
\usepackage{xcolor}         
\usepackage{graphicx}
\PassOptionsToPackage{square,numbers}{natbib}
\usepackage{amsmath}
\usepackage{amsfonts}
\usepackage{mdframed}

\usepackage{amssymb}
\usepackage{stackengine}
\usepackage{float}
\usepackage{wrapfig}
\usepackage{neurips_data_2024}
\usepackage{caption} 
\usepackage{graphicx}

\title{StableSemantics: A Synthetic Language-Vision Dataset of Semantic Representations in Naturalistic Images}

\begin{document}

\maketitle
\vspace{-1.0em}
\noindent\makebox[1.0\textwidth][c]{
\begin{minipage}{0.33\textwidth}
\begin{center}
  \textbf{Rushikesh Zawar*}\\
  Carnegie Mellon University\\
  rzawar@andrew.cmu.edu
  \vspace{2em}
\end{center}
\end{minipage}
\begin{minipage}{0.33\textwidth}
\begin{center}
  \textbf{Shaurya Dewan*}\\
  Carnegie Mellon University\\
  srdewan@andrew.cmu.edu
  \vspace{2em}
\end{center}
\end{minipage}
\begin{minipage}{0.33\textwidth}
\begin{center}
  \textbf{Andrew F. Luo}\\
  Carnegie Mellon University\\
  afluo@cmu.edu
   \vspace{2em}
\end{center}
\end{minipage}}\newline
\noindent\makebox[1.0\textwidth][c]{
\begin{minipage}{0.33\textwidth}
\begin{center}
  \textbf{Margaret M. Henderson}\\
  Carnegie Mellon University\\
  mmhender@cmu.edu
   \vspace{2em}
\end{center}
\end{minipage}
\begin{minipage}{0.33\textwidth}
\begin{center}
  \textbf{Michael J. Tarr} \\
  Carnegie Mellon University\\
  michaeltarr@cmu.edu
   \vspace{2em}
\end{center}
\end{minipage}
\begin{minipage}{0.33\textwidth}
\begin{center}
  \textbf{Leila Wehbe}\\
  Carnegie Mellon University\\
  lwehbe@cmu.edu
  \vspace{2em}
\end{center}
\end{minipage}}
\begin{abstract}
Understanding the semantics of visual scenes is a fundamental challenge in Computer Vision. A key aspect of this challenge is that objects sharing similar semantic meanings or functions can exhibit striking visual differences, making accurate identification and categorization difficult. Recent advancements in text-to-image frameworks have led to models that implicitly capture natural scene statistics. These frameworks account for the visual variability of objects, as well as complex object co-occurrences and sources of noise such as diverse lighting conditions. By leveraging large-scale datasets and cross-attention conditioning, these models generate detailed and contextually rich scene representations. This capability opens new avenues for improving object recognition and scene understanding in varied and challenging environments. Our work presents \textbf{StableSemantics}, a dataset comprising 224 thousand human-curated prompts, processed natural language captions, over 2 million synthetic images, and 10 million attention maps corresponding to individual noun chunks. We explicitly leverage human-generated prompts that correspond to visually interesting stable diffusion generations, provide 10 generations per phrase, and extract cross-attention maps for each image. We explore the semantic distribution of generated images, examine the distribution of objects within images, and benchmark captioning and open vocabulary segmentation methods on our data. To the best of our knowledge, we are the first to release a diffusion dataset with semantic attributions. We expect our proposed dataset to catalyze advances in visual semantic understanding and provide a foundation for developing more sophisticated and effective visual models. \url{https://stablesemantics.github.io/StableSemantics/}
\end{abstract}
\section{Introduction}
Visual scene understanding is a complex task that requires the integration of cues, context, and prior knowledge to navigate the inherent variability and complexity of the visual world. This complexity is particularly evident when considering the diversity of visual appearances that can correspond to a single semantic concept. For instance, entities that correspond to "man-made structures" can have vastly different visual appearances, ranging from sleek skyscrapers to rustic cottages. Similarly, objects that serve the same purpose, such as "containers," can have diverse shapes, sizes, and materials. This disconnect between semantic meaning and visual appearance poses a significant challenge for Computer Vision systems \cite{brust2018not, Duan2021BridgingGB, Alqasrawi2016BridgingTG, Barz2020ContentbasedIR}, requiring the disentanglement of the underlying semantic structure from visual differences \cite{caron2021emerging,xu2023open,elharrouss2021panoptic,vs2024possam,hu2023you, Quinn2017SemanticIR}. To overcome this challenge, recent advances in Computer Vision have adopted data-driven approaches, which learn to recognize patterns and relationships in large datasets of images and annotations. However, the reliance on large datasets of images and annotations poses a significant challenge in the development of Computer Vision systems. Acquiring and annotating such datasets can be a time-consuming and resource-intensive process, requiring careful consideration of data quality and diversity. 

This limitation has sparked interest in exploring alternative approaches that can reduce the need for large human-annotated datasets. One promising direction is the use of generative models, which have shown impressive results in translating between semantic meaning and visual appearance~\cite{rombach2022high,podell2023sdxl,song2020denoising,ho2022imagen}. In particular, diffusion-based text-to-image synthesis models have demonstrated an impressive ability to generate highly realistic images from textual descriptions, suggesting that these models must possess an implicit understanding of the semantic structure of the visual world, and have learned to associate words and phrases with specific visual concepts. By leveraging cross-attention mechanisms, these models learn to link textual input to visual representations and enable the generation of images that are grounded in the semantic content of the input text~\cite{tang2022daam}.
\begin{figure}[t!]
  \centering
\includegraphics[width=1.0\linewidth]{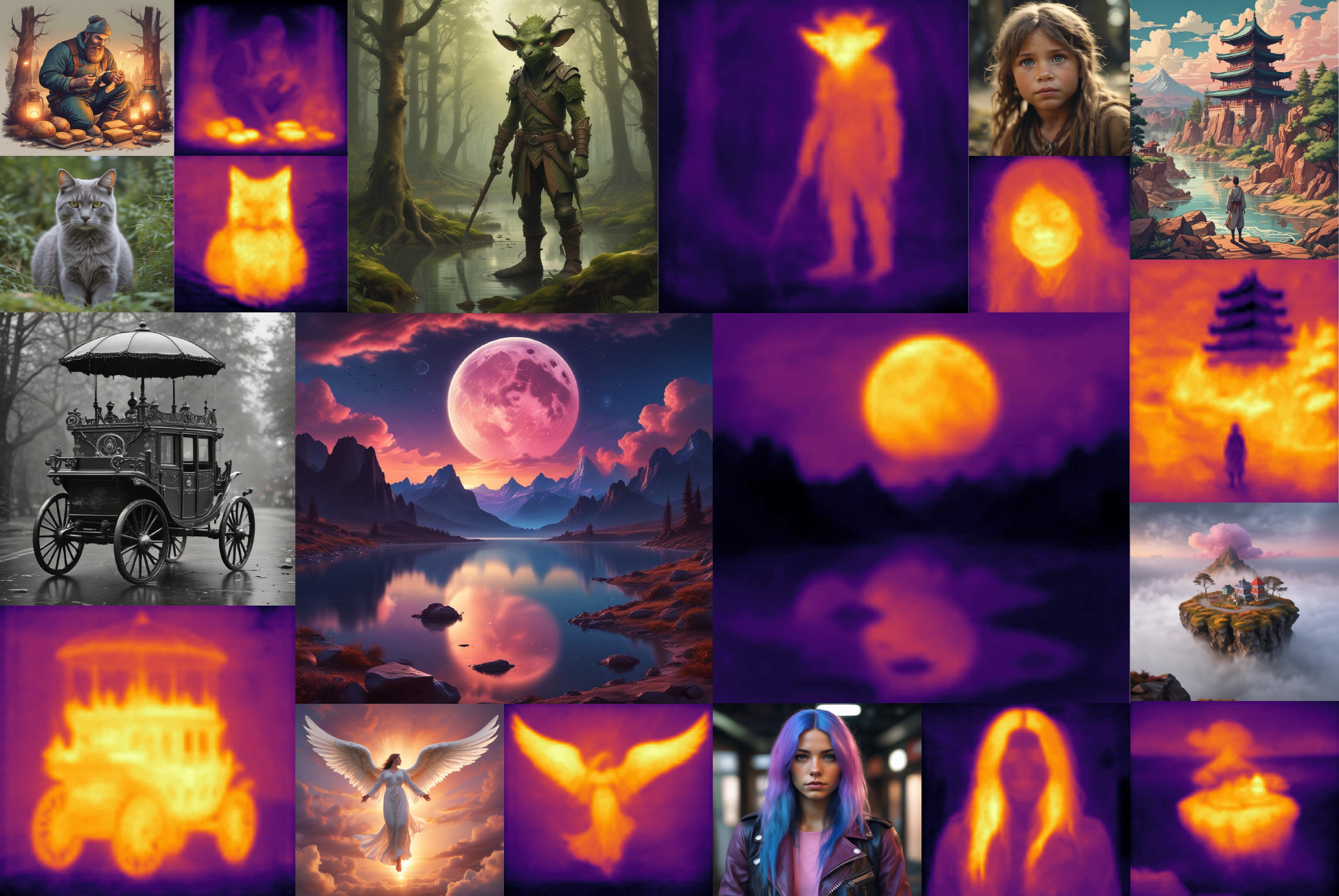}
  \caption{\textbf{Images and maps corresponding to select noun chunks from StableSemantics.} Images are generated using natural language captions derived from human generated and curated prompts. For reproducibility, seeds are recorded for each generation. Noun chunks are extracted by performing dependency parsing the natural language captions. Semantic maps corresponding to each noun chunk is computed using the cross-attention maps with the DAAM \cite{tang2022daam} method. Only a single attention map is shown here for each image, please see below for additional examples. Yellow indicates high relevance, black indicates low relevance.}
  \label{fig:teaser}
\end{figure}\unskip
In this work, we introduce \textbf{StableSemantics}, a dataset that consists of human-generated and curated prompts, natural language captions, images generated from the captions, and attention attribution maps corresponding to objects in the captions. Unlike prior work which sourced unfiltered human-generated prompts, we source our prompts from a pool of images that have been evaluated by humans for their visual appeal and interest, resulting in a dataset that mirrors the types of images people find engaging. As the original prompts may not always reflect natural language, we employ a large language model to paraphrase and refine them into fluent and natural-sounding captions, thereby bridging the gap between human-generated prompts filtered for visual appeal and naturalistic language. Each natural language caption is provided to a Stable Diffusion XL model to generate high-resolution and reproducible images. Finally, we explicitly record the dense text-to-image cross-attention maps used to condition the image generation process. We visualize the distribution of semantics across images, evaluate the spatial distribution of semantic classes within images, and evaluate the alignment of current captioning and open-set segmentation models on our dataset. To our knowledge, our dataset is the first to systematically record the spatial distribution of cross-attention activations corresponding to individual noun chunks.  We hope that \textbf{StableSemantics} will inspire future research on the visual distribution of semantic concepts and the development of more interpretable text-to-image synthesis models.
\vspace{-.2cm}
\section{Related work}
\myparagraph{Natural Scene Statistics.} Natural image statistics have been a long-standing area of research in computer vision and neuroscience. The human visual system is thought to be adapted to the statistical properties of natural images, which are characterized by complex dependencies between pixels \cite{Girshick2011, VanDerSchaaf1996}. The power law distribution of gradient magnitude statistics is thought to be a result of the hierarchical, self-similar structure of natural images, which arises from the presence of edges, textures, and other features at multiple scales. Understanding natural image statistics has important implications for image recognition tasks \cite{zoran2013natural, heiler2005natural, fang2012visualizing}, and has inspired the development of a range of algorithms and models that are tailored to the statistical properties of natural images \cite{mechrez2019maintaining, kleinlein2022sampling,hepburn2023disentangling,xiang2024research,hepburn2021relation,talbot2023tuned}. Other work has also explored the semantic structure of visual data, seeking to understand how higher-level categories and concepts are reflected in the statistical patterns present in images. This work has shown that different categories of images, such as scenes and objects, exhibit distinct statistical patterns \cite{Torralba2003, Henderson2023}. These semantic statistics have important implications for the development of models that can effectively represent and analyze visual data.

\vspace{-0mm}\myparagraph{Deep Image Generative Models.} Recent progress on generative models has enabled the generation of images, video, text, and audio \cite{rombach2022high,podell2023sdxl,song2020denoising,ho2022imagen,bao2024vidu,gupta2023photorealistic,touvron2023llama,touvron2023llama,evans2024long}. Models rely on a variety of different mathematical assumptions and architectures. Variational autoencoders \cite{kingma2013auto, luhman2023high, Harvey2021ConditionalIG, razavi2019generating, van2017neural} and flow-based models \cite{rezende2015variational, tong2023improving, Dinh2014NICENI, Kingma2018GlowGF}, while highly efficient, tend to produce lower-quality samples. Generative Adversarial Networks (GANs) \cite{goodfellow2014generative, karras2019style, arjovsky2017wasserstein, brock2018large, Kang2023ScalingUG, Mirza2014ConditionalGA,zawar2022detecting} can yield high-fidelity samples but may neglect modes in the data and can exhibit unstable training dynamics. Auto-regressive methods \cite{Huang2023NotAI, Parmar2018ImageT, Lee2022AutoregressiveIG, Huang2023TowardsAI, Ramesh2022HierarchicalTI}, although capable of producing high-quality samples, typically experience slow sampling. Recent progress in energy/score/diffusion models~\cite{rombach2022high,Ramesh2022HierarchicalTI} has given us methods that are simultaneously stable during training and yield high-quality samples. 

\myparagraph{Visual Datasets.} Deep learning models have achieved remarkable results by leveraging vast amounts of data. There has been a significant push to collect large-scale datasets. Earlier works such as LAION-5B \cite{schuhmann2022laionb}, Flickr Caption \cite{Young2014FromID} and YFCC100M \cite{10.1145/2812802} scrape real-world data of image-caption pairs from web sources. COCO \cite{Lin2014MicrosoftCC} goes a step further to also provide pixel-level segmentation masks on top of the image-caption pairs. \cite{Agrawal2015VQAVQ, Goyal2016MakingTV, Marino2019OKVQAAV, 8046084, vga} introduce datasets specifically for the task of VQA. Given the difficulty of collecting real data, recently there has been a shift towards synthetic datasets. StableRep~\cite{tian2024stablerep} also demonstrated the usefulness of Stable Diffusion images in training contrastive image models. Pick-a-Pic \cite{kirstain2023pickapic} provides a dataset of image-caption pairs where each sample contains a pair of diffusion-generated images and the human preference between those images. JourneyDB \cite{sun2023journeydb} and DiffusionDB \cite{wangDiffusionDBLargescalePrompt2022} are the closest works to ours that release large-scale datasets of synthetic image-caption pairs. 
\section{Data collection}\label{sec3}
\vspace{-.2cm}
In this section, we provide details on the collection and creation process of our dataset. Our data originates from human-generated and curated prompts submitted publicly by users online for Stable Diffusion XL. We describe our prompt collection process in section~\ref{promptcollect}. The prompts are filtered and transformed into natural language captions, and we describe our procedure in section~\ref{natlang}. Finally, we generate images and compute noun-chunk to image saliency maps via cross-attention in section~\ref{diffusiongen}.
\begin{figure}[t!]
  \centering
\includegraphics[width=1.0\linewidth]{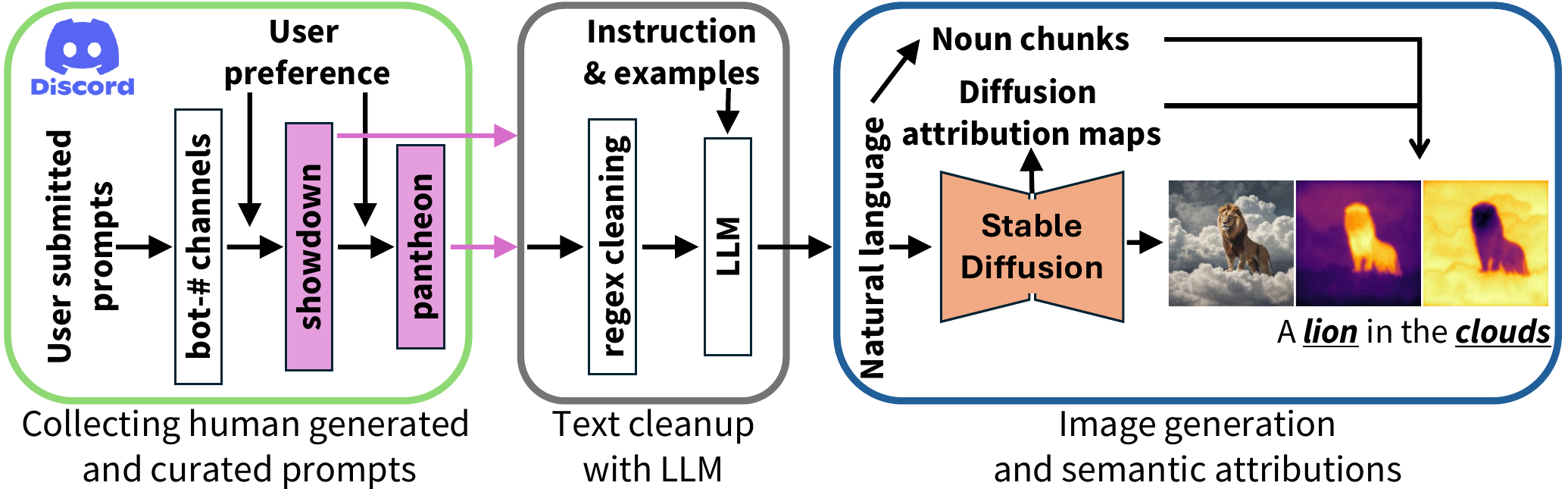}
   \vspace{-5mm}
  \caption{\textbf{Data collection and generation process}. \textbf{(1)} We collect our data from Stable Diffusion Discord, specifically the \textit{showdown} and \textit{pantheon} channels which are derived from user rankings of images generated from public prompt submissions. \textbf{(2)} The prompts are cleaned using regex to remove common errors, and further processed using an LLM to generate natural language captions. \textbf{(3)} The natural language captions are provided to a Stable Diffusion XL model, while we record the attention attribution maps corresponding to noun chunks.}
  \vspace{-0.5cm}
  \label{fig:arch}
\end{figure}
\subsection{Collecting human curated prompts}
\label{promptcollect}
Our dataset is collected from the \textbf{\textit{Stable Diffusion}} discord server, where users can publicly submit prompts to generate images using a discord bot. After users submitted prompts using the \texttt{/dream} command, the bot would return images corresponding to a prompt. Beyond accepting a prompt, users could also submit negative prompts, and image styles which were achieved via a prefix/affix pattern of text to the original prompt. These style patterns were not visible to the users. 

We started our data collection after the Stable Diffusion XL 1.0~\cite{podell2023sdxl}  candidate was made available via bots. The data was continuously collected from \textbf{July 11, 2023} (a day after SDXL 1.0 candidate bots were launched) until \textbf{Feb 07, 2024} (SDXL bot shutdown). Users were allowed to submit prompts to \texttt{bot-\#} channels where \texttt{\#} corresponds to a number. We observed that the number of channels varied over time, and generally remained at slightly over $10$. For each prompt, the bot would return $2$ images. Users were asked to select which image was better by clicking on a button corresponding to an image, without explicit guidance on what "better" meant. Our understanding from discussions with members of staff was that these prompts and image pairs were used for fine-tuning the SDXL candidates using RLHF/DPO ~\cite{ouyang2022training,rafailov2024direct}, selection of model candidates, and selection of generation hyperparameters.

Prior work has also collected user-generated prompts from discord servers for MidJourney and Stable Diffusion~\cite{sun2023journeydb,wangDiffusionDBLargescalePrompt2022}. Our work goes further by only collecting prompts that were human-curated. The Stable Diffusion discord followed a three-tier hierarchy for prompts, where users first submit and rate images in \texttt{bot-\#}, with highly rated images from all bot channels going into a single \texttt{showdown} channel every $15$ minutes. The \texttt{showdown} channel was reset every $30$ minutes and had the history wiped. In the showdown channel, 2 images and their respective prompts were placed side-by-side. Users again were asked to select the images that were more visually appealing. Every $30$ minutes, the top-ranked images and prompts would go into the \texttt{pantheon} channel. The \texttt{pantheon} had history going back to inception May 02, 2023. We note that strictly speaking the \texttt{showdown} to \texttt{pantheon} selection process was not fair to images that came in at the second $15$ minute slice, as they were given less time to be voted upon. Due to this, we do not further distinguish between prompts collected from these two sources. Our data collection process ran every $14$ minutes on the \texttt{showdown} channels and ran once on the \texttt{pantheon} channel. This was sufficient as after the initial collection date, new \texttt{pantheon} entries were a strict subset of \texttt{showdown} prompts. Visual inspection of the generations from \texttt{showdown} and \texttt{pantheon} suggest that these images were generally more artistic and contained more interesting visual compositions than the \texttt{bot-\#} channel. We collect a total of 235k unique user-generated prompts, which is further filtered according to NSFW ratings and caption length. 
\subsection{Obtaining natural language captions}
\label{natlang}
As shown in Figure~\ref{dsetexamples}, the user-submitted prompts generally took a tag-like format, with descriptors being separated by commas. Such prompts are convenient for users to specify and likely achieve good results due to the use of CLIP text networks for conditioning, which can operate like bag-of-words models~\cite{thrush2022winoground,yuksekgonul2022and}. However, such prompts generally perform poorly when typical NLP pipelines are used for analysis. These prompts further may not explicitly specify needed visual relationships in the text, and instead excessively rely on the prior learned by the diffusion model to disambiguate relationships.
In order to mitigate this issue, we utilize an LLM model to clean up the original raw user-generated prompts. We use \texttt{Gemini 1.0 Pro} for this task, as it performed competitively against other models at the time of our work~\cite{team2023gemini} and offered a free API. The model was instructed to take the user-generated prompts and transform them into natural language captions. To enhance the results, we augment the prompt via in-context learning from GPT-4 input/output pairs. To remove NSFW prompts, we record the Gemini API safety ratings for each input prompt, and remove the prompts where a 4 out of 4 rating was given on the axes of sexuality/hate speech/harassment, or if the model itself produced a refusal, or if the prompt was repeatedly returned with an error (blocked by Google). Please see Figure~\ref{fig:arch} for a visualization of the pipeline.
\vspace{-.2cm}
\subsection{Image generation and semantic attribution}
\begin{table}
\centering
\resizebox{1.0\linewidth}{!}{
\centering
\begin{tabular}{cccccc}
\toprule \addstackgap{Dataset}         & Total Images & Total Captions & Caption Source & Human Preferred & Open-set Semantics \\ \hline
COCO 2017~\cite{Lin2014MicrosoftCC}       & \addstackgap{123k}         & 617k           & H              & N               & N                  \\
LAION-COCO~\cite{schuhmann2022laionb}     & 600M         & 600M           & M              & N               & N                  \\
DiffusionDB~\cite{wangDiffusionDBLargescalePrompt2022}    & 14M          & 1.8M           & H              & N               & N                  \\
JourneyDB~\cite{sun2023journeydb}      & 4.7M         & 1.7M           & H+M            & N               & N                  \\
\textbf{StableSemantics} & 2M           & 224k           & H+M            & Y               & 10.8M\\
\bottomrule
\end{tabular}}
\caption{\textbf{Size of the different components of StableSemantics.} Our captions are selected by humans to correspond to visually interesting images. We are the only dataset to provide dense open-set spatial semantic maps. Our maps are derived from the cross-attention maps in Stable prompts. Note that 235k unique captions are collected, 224k remain after NSFW filtering and only 200k captions are used for image generation after filtering for length.}
\end{table}\unskip
\begin{figure}[t!]
    \centering
        \centering
        \includegraphics[width=1.0\textwidth]{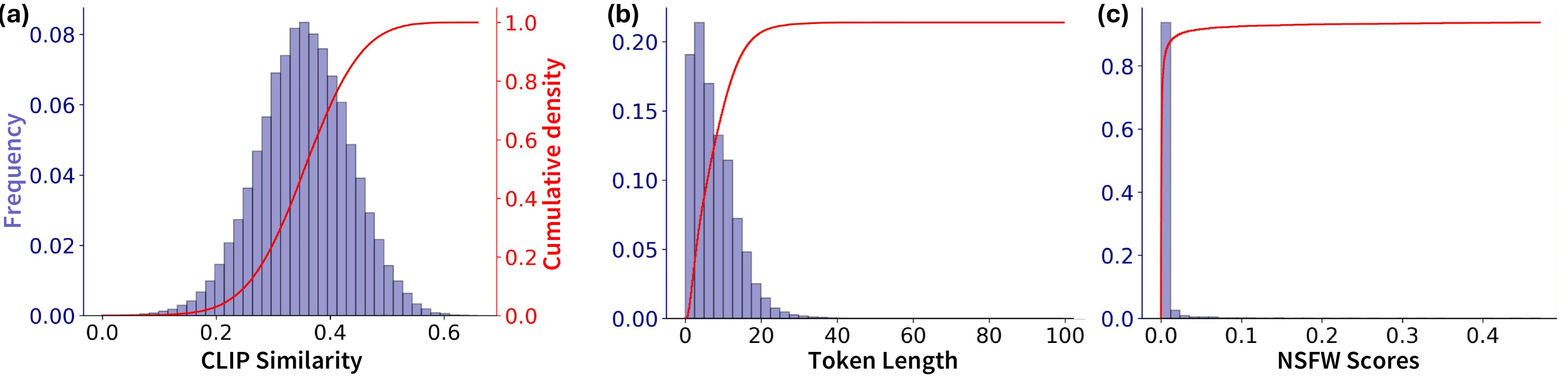}
    \caption{\textbf{Histogram of dataset statistics.} \textbf{(a)} We visualize the cosine CLIP similarities between generated images and original captions. \textbf{(b)} Number of tokens in the captions. \textbf{(c)} NSFW scores of the captions after LLM filtering. Scores measured by LLaMA Guard 2 for sexuality and hate.}
\label{histo}
\end{figure}\unskip
\label{diffusiongen}
To provide a fully reproducible pipeline for the images and maximize the usability of our dataset, we generate the images ourselves using open weights and record the random seed for each generation. Images are generated using \texttt{sdxl lightning 4step unet}~\cite{lin2024sdxl}, a few-step distilled version of Stable Diffusion XL. For each prompt, we perform parsing using spaCy \texttt{en\_core\_web\_lg} to extract noun chunks. To obtain mappings from noun chunks to spatial attributions, we use Diffusion Attentive Attribution Maps which measures the cross-attention from tokens in the language condition to the UNet. Specifically, we used the improved DAAM-i2i guided heatmap variant~\cite{tang2022daam,daami2i} which improves object localization. We observe that unrelated articles like "a", and "the" and possessive determiners like "his", "her", "our", "their" are not typically localized to a specific object, but rather have attribution maps diffuse over the background or various objects in the image. While similar phenomena has been noted in ViTs~\cite{darcet2023vision} and pure text LLMs~\cite{clark2019does,kovaleva2019revealing,xiao2023efficient}, our observation is novel in that text-to-image diffusion models are encoding contextual information in these "filler" words. For effective localization, we remove articles and possessive determiners if they are the first word of a noun chunk.

\vspace{-.2cm}
\section{Dataset exploration}
\label{explore}
\vspace{-.2cm}
\begin{figure}[t!]
  \centering
\includegraphics[width=1.0\linewidth]{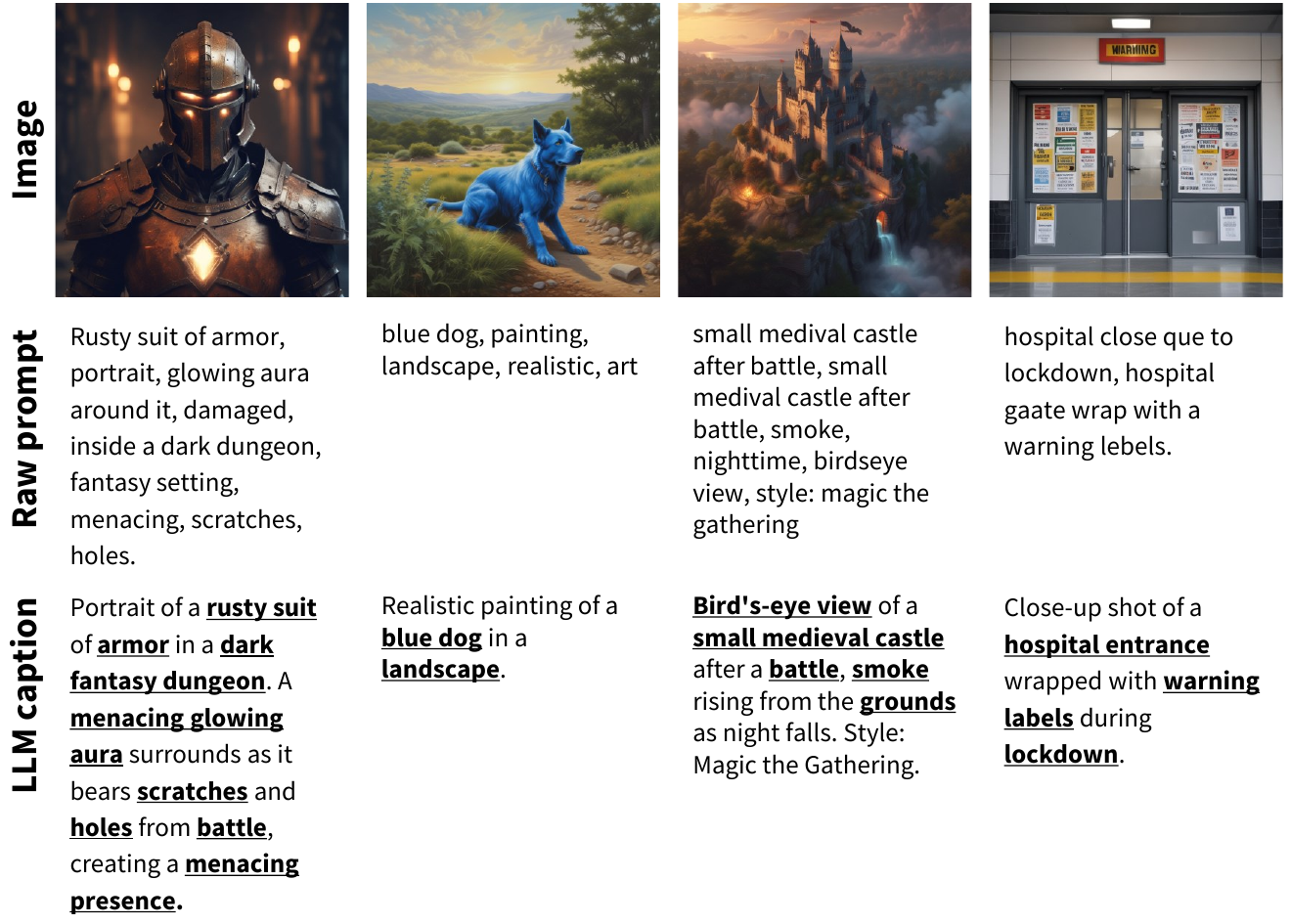}
   \vspace{-1mm}
  \caption{\textbf{Example of SDXL generated images from the captions, raw user prompts and LLM processed captions.} Raw prompts from users often contain typos or take the form a non-natural language tag-like format. We instruct an LLM to transform the prompts into a natural language caption. \underline{\textbf{Noun chunks}} (bolded and underlined) are derived from dependency parsing. Images are generated from the captions, with diffusion attribution maps recorded for the noun chunks. }
  \vspace{-0.2cm}
  \label{dsetexamples}
\end{figure}\unskip
 We first evaluate the CLIP similarity between the generated images and the captions, and further characterize the safety and length of the captions. We then explore the semantic distribution of the images and the captions using CLIP, and visualize the spatial distribution of objects in a scene using diffusion attribution maps. Finally, we evaluate the performance of captioning and open-set segmentation models on our dataset. These characterizations demonstrate how StableSemantics can be a promising dataset for advancing visual semantic understanding. The data will be released under a CC0 1.0 license.

\subsection{Dataset characterization}
 After deduplication and LLM NSFW filtering, we have 224 thousand natural language captions. We evaluate the similarity of the SDXL-lightning generated images and the captions in Figure~\ref{histo}a using OpenAI's CLIP ViT-B/16~\cite{Radford2021LearningTV}. We find that the CLIP similarity peaks at $0.34$, which is similar to CLIP scores achieved using SDXL. These scores typically range from $0.2$ to $0.5$ which means that our prompts and images can be interpreted to be semantically very similar given the higher range of scores. We visualize the token length of the captions in Figure~\ref{histo}b. Note that we do not generate images for captions exceeding $77$ tokens post-padding. This yields a total of 200k captions which are used for image generation. In Figure~\ref{histo}c, we plot the NSFW scores of the captions used for image generation, as evaluated using the state-of-the-art \texttt{Meta Llama Guard 2} model. We define the unsafe categories to the 3 official categories relating to sexual content, and the 1 official category related to hate speech. The scores are the "unsafe" softmax outputs between the "safe" and "unsafe" tokens. We find that the captions used for generation are overwhelmingly safe. In Figure~\ref{dsetexamples}, we provide examples of the images, the original human generated prompts which may often contain typos or tags, and the LLM output natural language prompts. Likely due to human preference, we observe a higher ratio of images with visually interesting compositions.

\begin{figure}[t!]
  \centering
\includegraphics[width=1.0\linewidth]{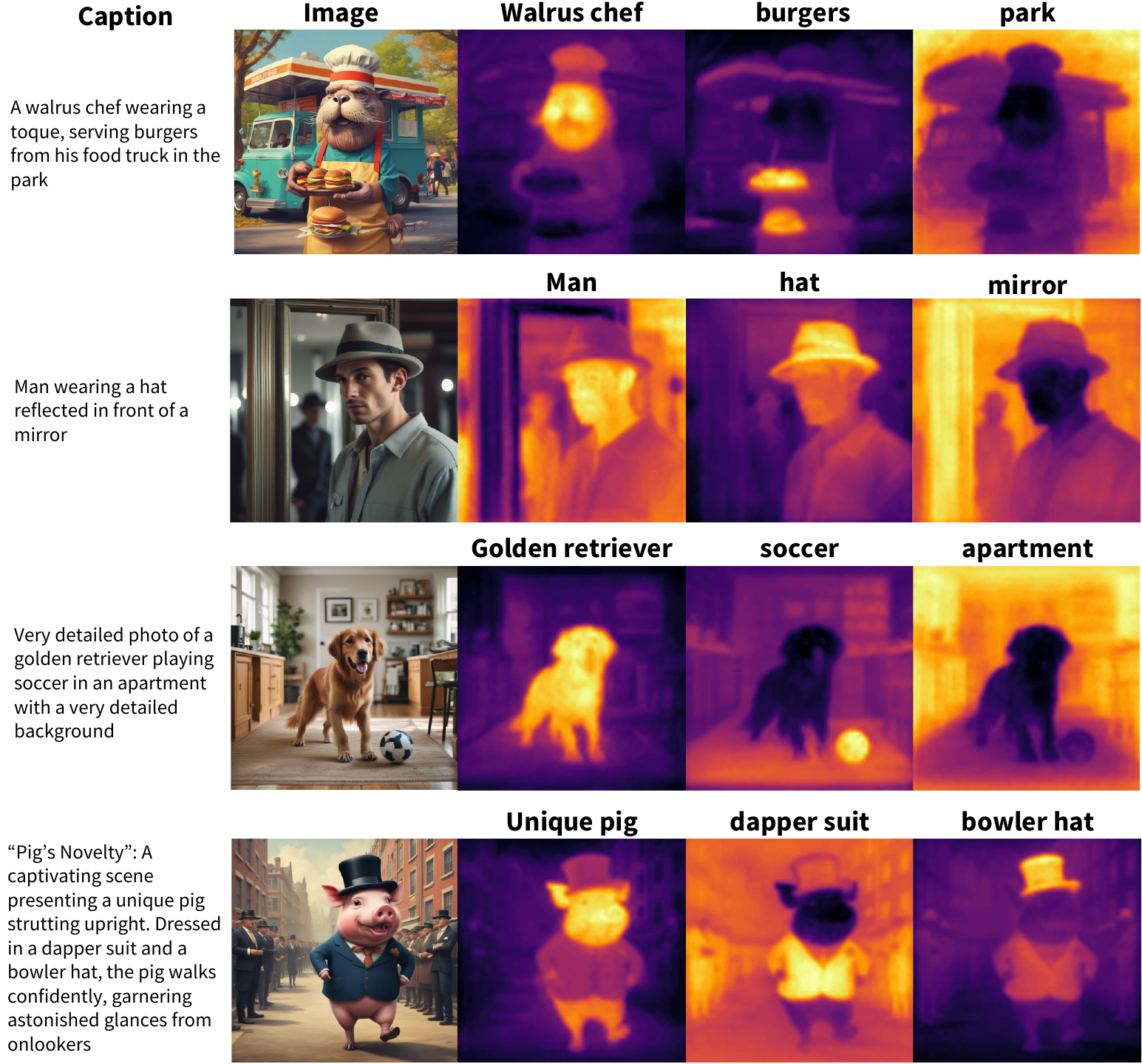}
  \caption{\textbf{Visualization of the dataset.} We show example captions used for image generation, images generated from the captions, and select noun chunks and their corresponding attention attribution maps. We find that our dataset contains accurate localizations for different semantic concepts.}
  \label{semanticmasks}
\end{figure}\unskip
\begin{figure}[h!]
  \centering
\includegraphics[width=1.0\linewidth]{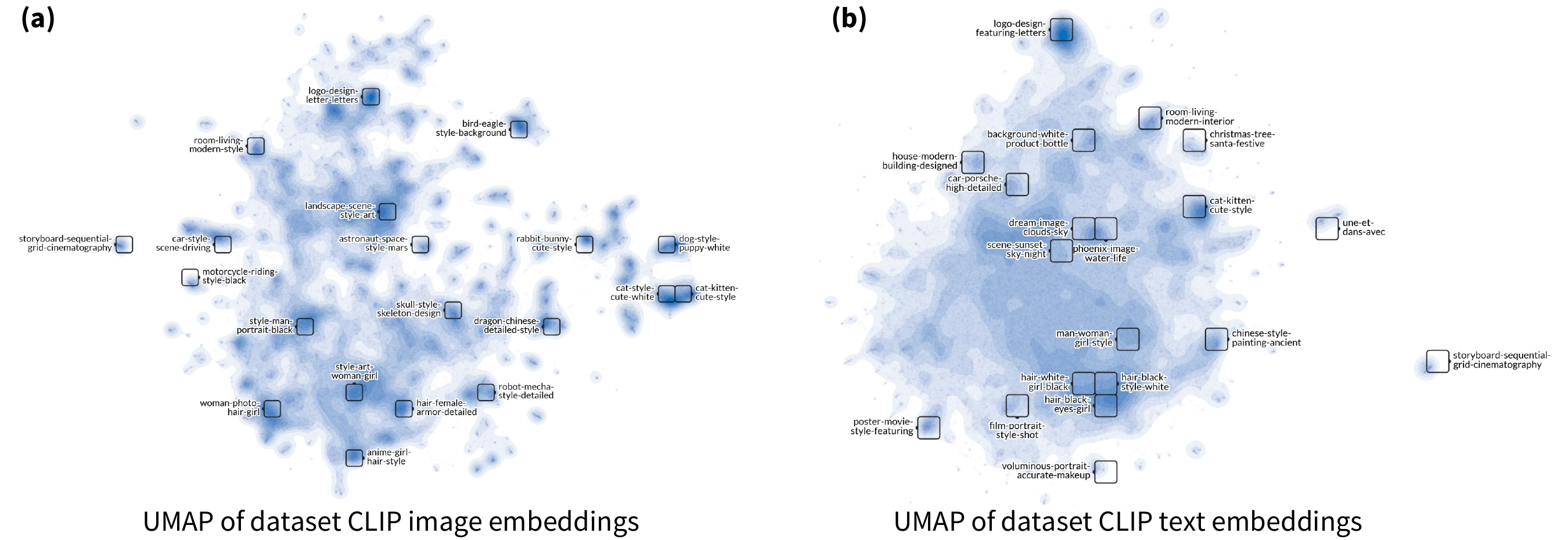}
   \vspace{-1.5mm}
  \caption{\textbf{UMAP visualization of dataset CLIP embeddings.} We use OpenAI CLIP ViT-B/16 to compute embeddings for both the generated \textbf{(a)} images and the \textbf{(b)} text. UMAP with the cosine metric is used to perform dimensionality reduction. We observe that images describing people, scenes, text, and animals occur with high frequency.}
  \vspace{-0.25cm}
  \label{umap}
\end{figure}\unskip
\subsection{Semantic exploration of the dataset}
In Figure~\ref{umap} we visualize the semantic distribution of whole images and the captions used to generate the images. We utilize UMAP~\cite{mcinnes2018umap} with a cosine metric applied to CLIP embeddings for this visualization with wizmap~\cite{wang2023wizmap}. We find that the distribution of both images and text exhibit peaks in concepts such as people, scenes, text, and animals (cats and dogs). These peaks likely reflect the effect of human preference on visually interesting images.

The semantic maps we provide in our dataset help localize image regions corresponding to specific noun chunks from the prompts. In Figure~\ref{semanticmasks} we visualize the captions used from image generation, the generated RGB image, and attention attribution masks corresponding to noun chunks shown in bold. We find that our dataset can provide semantic attributions that are well aligned to objects in the scene. This is likely due to the nature of Stable Diffusion, which leverages cross-attention guidance to generate complex compositional images.
\begin{figure}[t!]
  \centering
\includegraphics[width=1.0\linewidth]{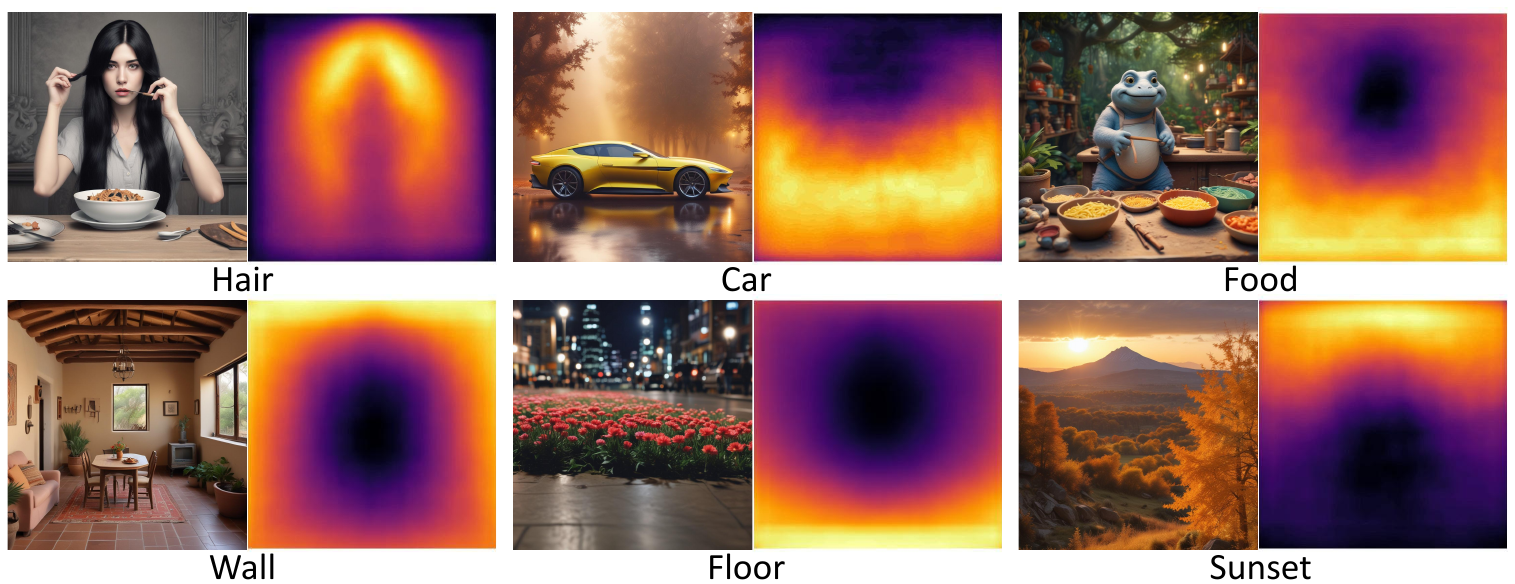}
  \caption{\textbf{Spatial distribution of semantic concepts.} For each concept, we visualize an example image containing the concept, as well as the spatial distribution averaged over occurrences. We utilize CLIP text similarity to select the top-100 most similar noun chunks and average those occurrences.}
  \vspace{-.5cm}
  \label{fig7}
\end{figure}\unskip 
We use these maps to analyze whether our images exhibit a trend of certain concepts being generated in specific regions on average. In Figure~\ref{fig7} we aggregate the masks of the top 100 noun chunks that have the highest CLIP similarity scores with concepts of interest. We apply this similarity based matching to allow for inexact matching. Figure~\ref{fig7} clearly shows that the spatial distribution of concepts can be highly non-uniform. This bias likely reflects the distribution of concepts in natural images~\cite{Torralba2003, Greene2013}. For instance, it makes sense for the sunset to always be on the top, the walls towards the sides, and the floor towards the bottom. It is also very common to find human beings as the primary subjects in images which explains the placement of hair surrounding a central region. Finally, many images exhibit food on top of a table and cars on roads. In these scenarios, these semantic concepts typically occupy the bottom half of the visual field. We visualize \cite{d3js} the frequency distribution of nouns grouped by wordnet hierarchy in Figure~\ref{hier}. Our dataset could be used to understand the spatial and visual bias present in natural images.
\begin{figure}[t!]
  \centering

\includegraphics[width=1.0\linewidth]{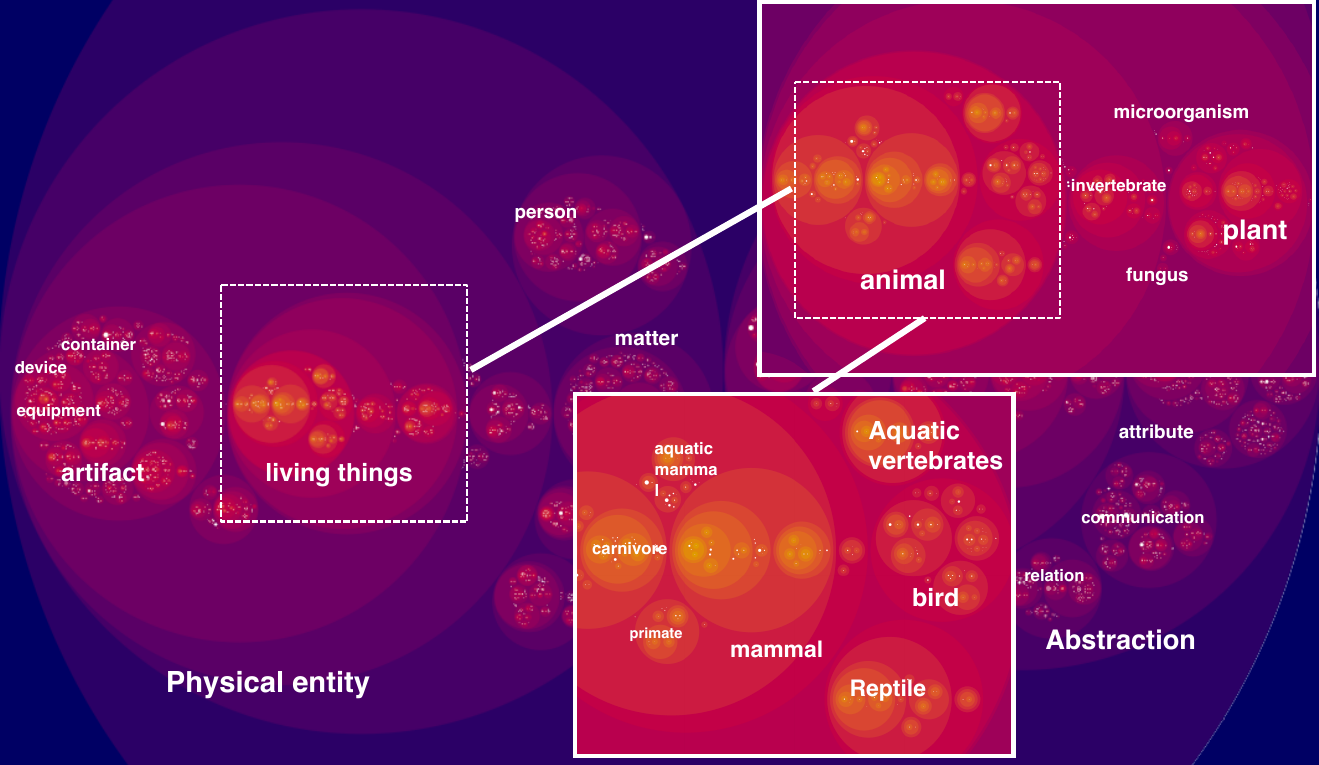}
   \vspace{-1mm}
  \caption{\textbf{Frequency of nouns visualized with wordnet hierarchy.} We parse the sentences and extract the nouns. The hierarchy is from wordnet~\cite{fellbaum2010wordnet}. The circle size corresponds to frequency.}
  \label{hier}
\end{figure}\unskip
\subsection{Evaluation of models}
In this section we evaluate the performance of state-of-the-art open-vocabulary image segmentation and captioning models on our dataset. 
For open-vocabulary segmentation methods, we evaluate the standard mean Intersection over Union (mIOU), where discrete masks are computed by taking the argmax over all noun chunks' continuous masks for a given prompt. As these methods also produce soft masks, we also evaluate the pearson correlation of the attribution maps from our datasets. In Table~\ref{tab:1}, We find that recent open-vocabulary segmentation models which modify CLIP (LSeg, SCLIP) or leverage text-to-image diffusion models (ODISE) perform better than their peers.

We also evaluate recent vision-language models for image captioning in Table~\ref{tab:2}. We evaluate the generated captions against the original captions using state-of-the-art E5-Mistral model to evaluate cosine similarity ($\times 100$ for clarity), BLEU-4, and CIDEr scores. These results suggest that while the captions predicted by models may use different wording from the original caption, they can be semantically very similar.

\begin{table}[h]
\centering
\makebox[0pt][c]{\parbox{1.0\textwidth}{%
    \begin{minipage}[t]{0.48\hsize}\centering
\begin{tabular}{@{}lcc@{}}
\toprule
     \textbf{Method} & \textbf{mIoU $\uparrow$} & \textbf{Pearson $\uparrow$} \\
\midrule
    MaskCLIP~\cite{dong2023maskclip} & 0.015 & 0.199 \\
    SCLIP~\cite{wang2023sclip} & 0.109 & 0.236 \\
    LSeg~\cite{li2022language} & \textbf{0.164} & 0.032 \\
    CLIPSeg \cite{lueddecke22_cvpr} & 0.133 & 0.143 \\
    ODISE \cite{xu2023open} & 0.096 & \textbf{0.300} \\
    OVSeg \cite{liang2023open} & 0.035 & 0.181 \\
\bottomrule
\end{tabular}
    \caption{\textbf{Performance comparison of open-set segmentation models.} We evaluate the intersection over union and pearson correlation for noun chunks against model outputs.}
    \label{tab:1}
        \label{tab:singlebest}
    \end{minipage}
    \hfill
    \begin{minipage}[t]{0.48\hsize}\centering
\scalebox{0.75}{\centering\begin{tabular}{@{}lccc@{}}
\toprule
     \textbf{Methods} & \textbf{E5-Mistral\cite{wang2023improving} $\uparrow$} & \textbf{BLEU \cite{papineni-etal-2002-bleu} $\uparrow$} & \textbf{CIDEr \cite{vedantam2015cider} $\uparrow$} \\
\midrule
    LLaVA~\cite{liu2024visual} & 67.9 & 1.2  & 3.1 \\
    BLIP-2~\cite{li2023blip} & \textbf{70.9} & \textbf{1.9} & \textbf{10.2} \\
    GIT~\cite{wang2022git} & 63.3 & 1.0 & 6.8 \\
    CoCa~\cite{yu2022coca} & 66.8 & 1.7 & 9.7 \\
\bottomrule
\end{tabular}}
\vspace{1.6em}
\caption{\textbf{Performance comparison of captioning models.} We apply captioning models and evaluate the alignment of the outputs against the captions used to generate the images.}
\label{tab:2}

    \end{minipage}
    \hfill
}}
\end{table}
\vspace{-.2cm}
\section{Discussion}
\textbf{Limitations and Future Work.} Our work relies on human-submitted prompts, which may exhibit non-natural semantic co-occurrences. During the data collection process, we also observed a strong shift in the semantic distribution of prompts and images around holidays (Thanksgiving, Christmas). This suggests that continual data collection is required to mitigate bias. \\
\textbf{Conclusion.} We introduce StableSemantics, the first large-scale dataset that combines natural language captions, synthetic images, and diffusion attribution maps. Our work goes beyond prior datasets by providing spatially localized noun chunk to image region mappings. We explore the semantic distribution of whole images and objects within an image. The availability of this dataset will allow for the use of synthetic visual data in additional domains.
\clearpage
\bibliography{myref}
\clearpage
\renewcommand\thefigure{S.\arabic{figure}}    
\setcounter{figure}{0}
\renewcommand\thetable{S.\arabic{table}}   
\setcounter{table}{0}
\setcounter{section}{0}
\renewcommand\thesection{A.\arabic{section}}   

{\begin{center}\textbf{\Large Supplementary Material: \\StableSemantics: A Synthetic Language-Vision Dataset of Semantic Representations in Naturalistic Images}\end{center}}

\textbf{\Large Sections}

\begin{enumerate}
    \item Additional dataset visualization (section~\ref{visu})
     \item Visualization of object distributions  (section~\ref{supp_dists})
    \item Comparison of open vocabulary segmentation methods (section~\ref{supp_openseg})
    \item Prompt used for language model cleanup of raw prompts (section~\ref{prompt})
    \item Dataset documentation (section~\ref{docu})
    \item Broader impacts (section~\ref{impact})
    \item Dataset website (section~\ref{website})
    \item Statement of responsibility (section~\ref{responsibility})
    \item Dataset license (section~\ref{license})
    \item Hosting plan and dataset format (section~\ref{hosting})

\end{enumerate}
\clearpage
\section{Additional dataset visualization}
\label{visu}
\begin{figure}[ht]
  \centering
\includegraphics[width=1.0\linewidth]{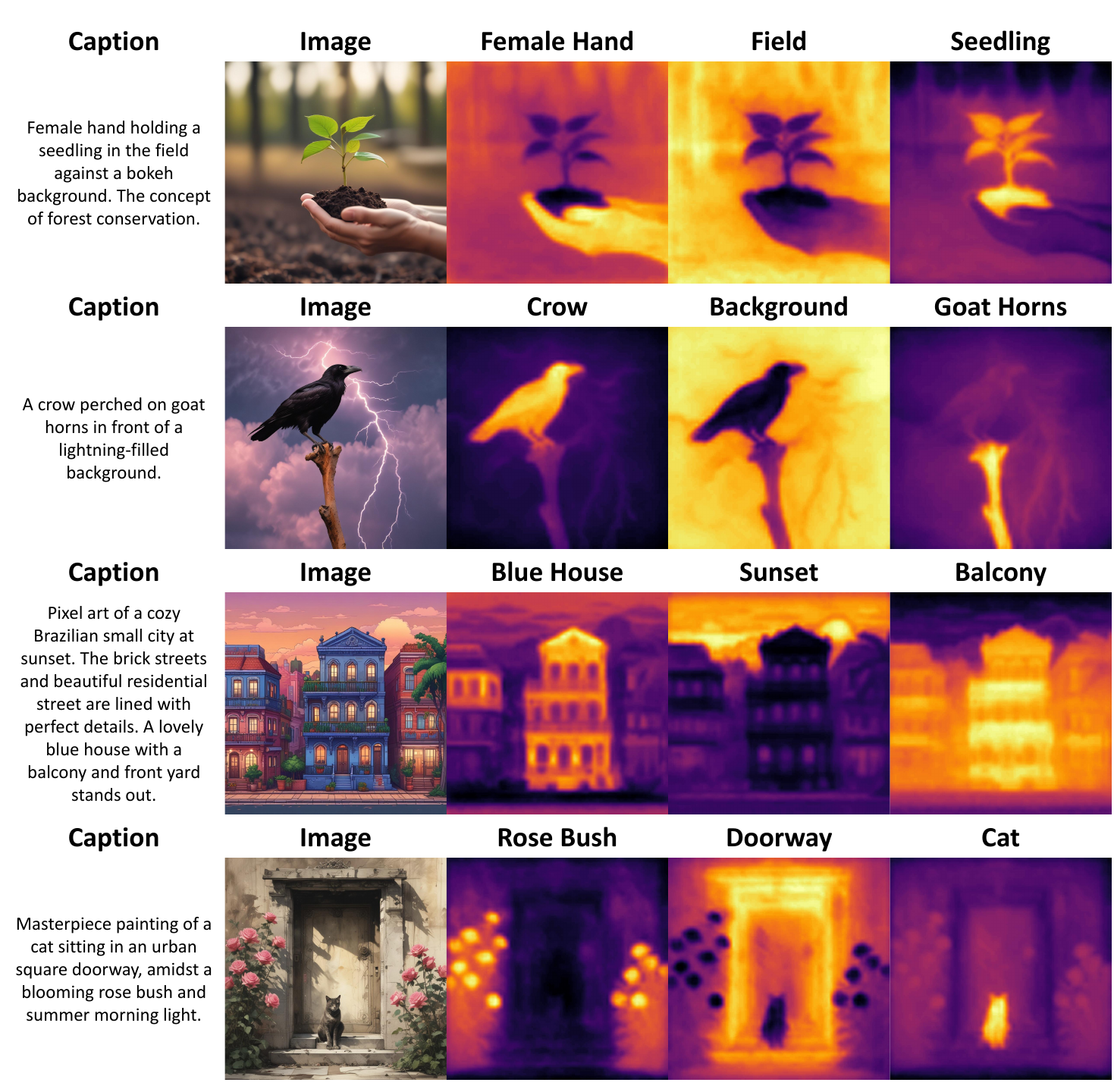}
  \caption{\textbf{Visualization of additional dataset examples.} We show the natural language caption used for the image generation, the image, and masks corresponding to select noun chunks.}
  \label{additional_masks}
\end{figure}

\begin{figure}[ht]
  \centering
  \vspace{-2em}
\includegraphics[width=1.0\linewidth]{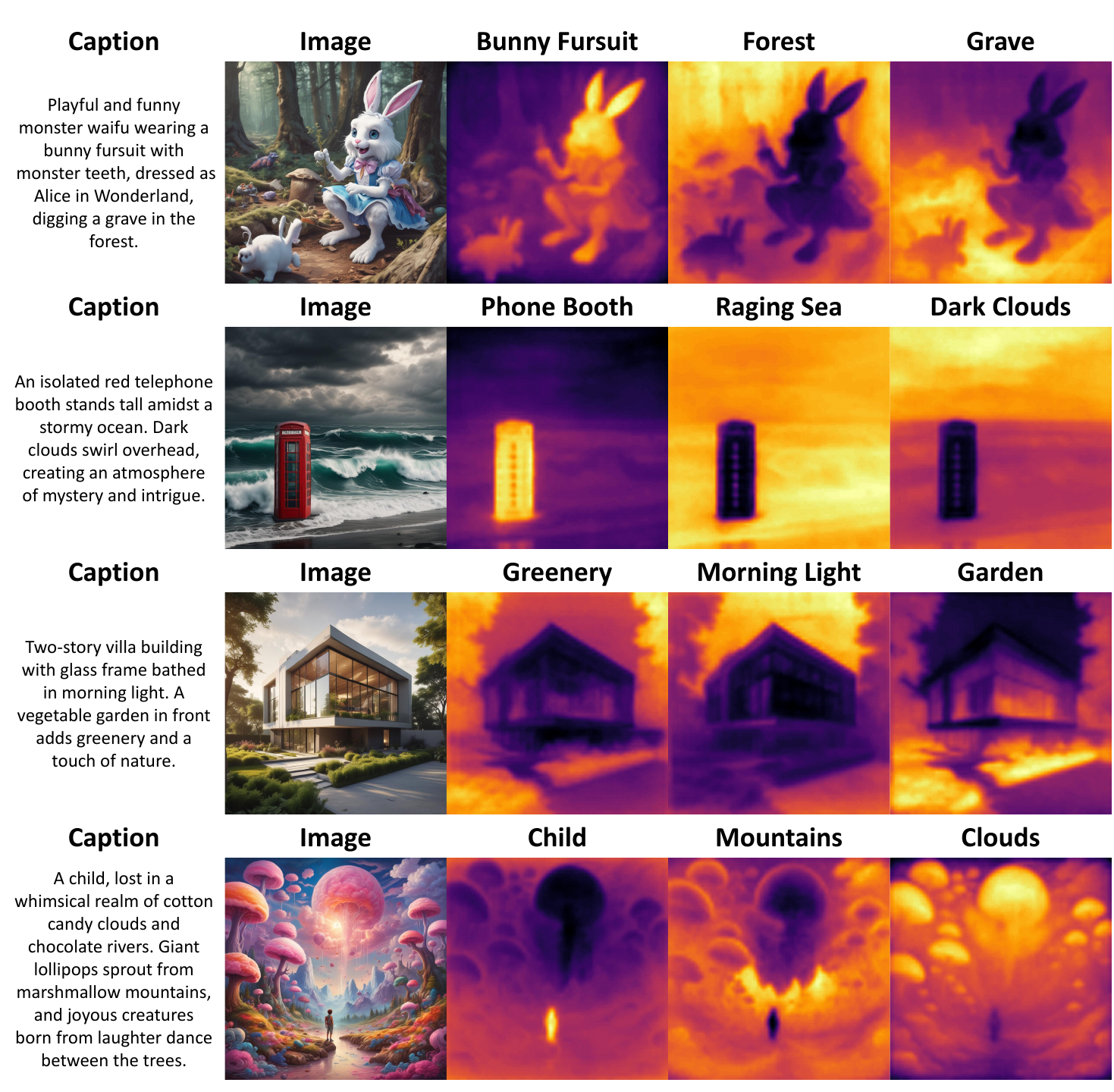}
  \caption{\textbf{Visualization of additional dataset examples.} We show the natural language caption used for the image generation, the image, and masks corresponding to select noun chunks.}
  \label{additional_masks2}
\end{figure}

In this section, we provide additional visualizations of the natural language captions generated using a large language model from the raw user prompts, the RGB image, and semantic masks corresponding to select noun chunks in the image in Figure~\ref{additional_masks} and Figure~\ref{additional_masks2}. 
\clearpage
\section{Visualization of object distributions}
\label{supp_dists}
\begin{figure}[ht]
  \centering
\includegraphics[width=1.0\linewidth]{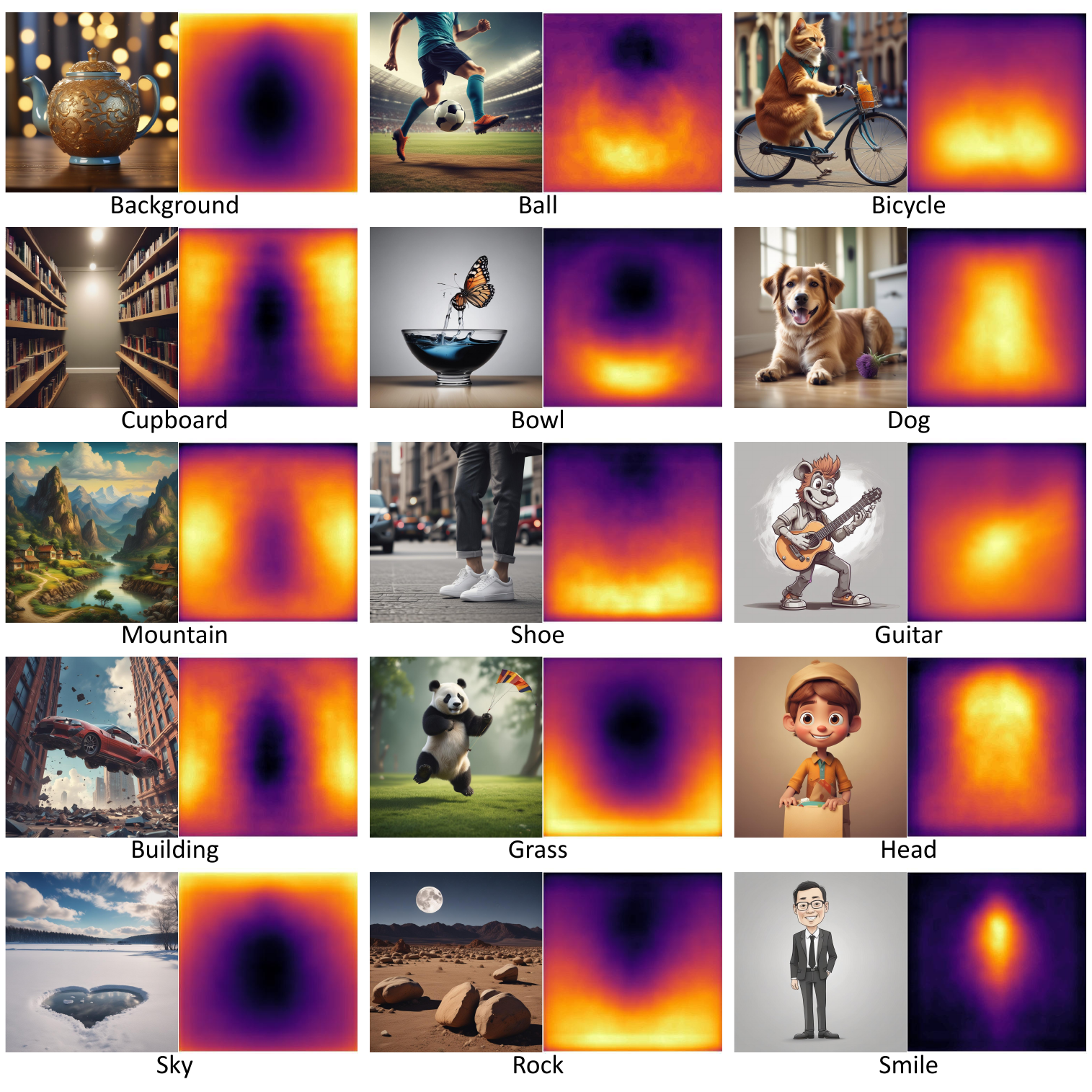}
  \caption{\textbf{Additional examples on the spatial distribution of concepts.} We provide additional examples of images containing a concept, and the average distribution of the top-100 images containing the most similar noun chunks as evaluated using the CLIP text model.}
  \label{fig:supp_avg}
\end{figure}

In this section, we visualize the spatial distribution of various noun chunks in Figure~\ref{fig:supp_avg}. We note that several types of objects exhibit highly non-uniform spatial distributions.
\clearpage
\section{Comparison of open vocabulary segmentation methods}
\label{supp_openseg}
\begin{figure}[ht]
  \centering
  \vspace{-2em}
\includegraphics[width=0.92\linewidth]{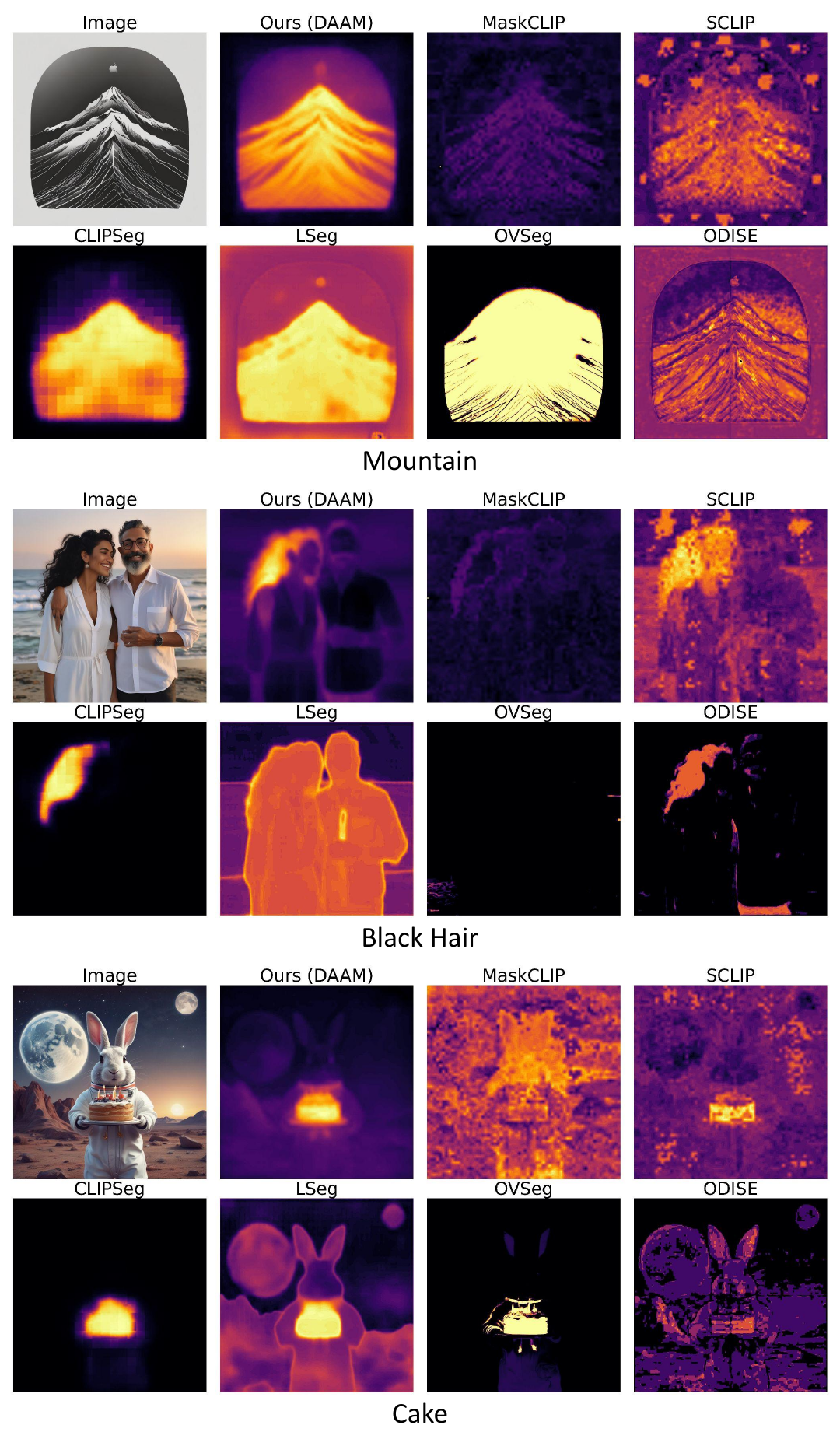}
  \caption{\textbf{Comparison of semantic maps and open vocabulary segmentation methods.} We visualize the semantic attribution maps corresponding to noun chunks from our dataset, and the segmentation maps produced by various state-of-the-art methods.}
  \label{supp_seg1}
\end{figure}

\begin{figure}[ht]
  \centering
\includegraphics[width=0.92\linewidth]{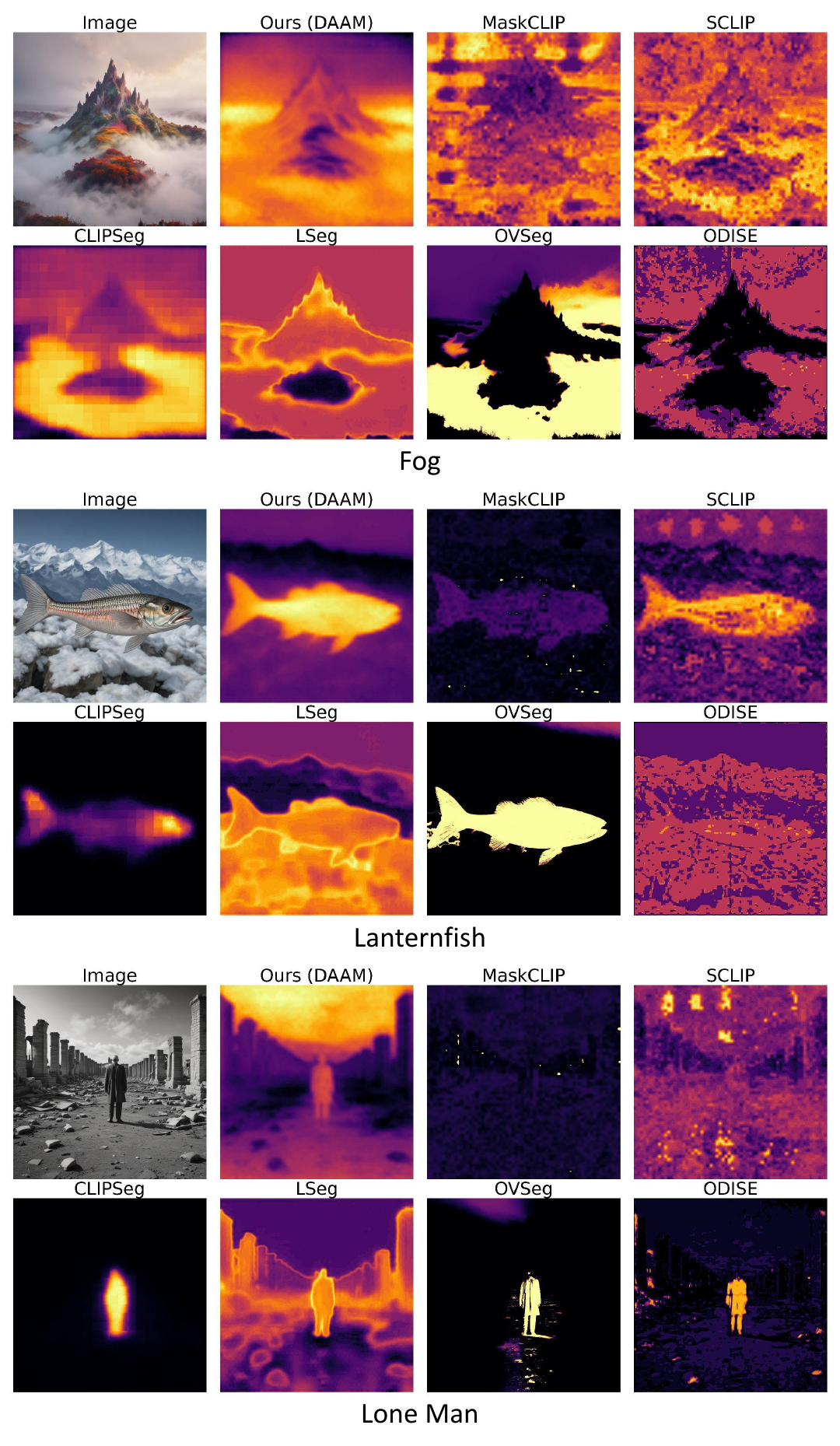}
   \vspace{-4mm}
  \caption{\textbf{Comparison of semantic maps and open vocabulary segmentation methods.} We visualize the semantic attribution maps corresponding to noun chunks from our dataset, and the segmentation maps produced by various state-of-the-art methods. Note that the \textbf{Lone Man} illustrates how the semantic attribution maps can be imperfect. In this case it includes additional background.}
  \label{supp_seg2}
\end{figure}

In this section, we provide additional visualizations of the semantic maps from our dataset, and segmentation outputs in Figure~\ref{supp_seg1} and Figure~\ref{supp_seg2}. We note that in general, the semantic masks in our dataset can accurately localize objects. However, there are specific cases (Lone Man) where the attention maps corresponding to noun chunks can include other contextual objects. We believe this occurs when the diffusion model tries to generate co-occurring scene and image parts that are not explicitly mentioned in the caption.
\clearpage
\section{Prompt used for language model cleanup of raw prompts}
\label{prompt}
We utilize the following prompt followed by the raw user prompt to obtain a natural language caption. Note that our raw prompts undergo simple \texttt{regex} based processing to remove some obvious errors before being provided to the language model.\\
\\
\texttt{You are going to be provided with the description of an image. You will transform and edit the description as needed into a cohesive natural language sentence or sentences without elaborating. If the original is mixed language, then your output should also be mixed language. Do not elaborate, do not provide information about people mentioned in the description.\\ \\Make a best effort to use ALL WORDS AND DETAILS from the original description. DO NOT MAKE UP DETAILS unless absolutely necessary. BE AS CONCISE AS POSSIBLE, WHILE ATTEMPTING TO INCLUDE ALL WORDS AND DETAILS FROM THE ORIGINAL DESCRIPTION. You may only omit details or words if they are nonsensical or form a contradiction. Attempt to fix typos and remove invalid punctuation. Omit emojis in your output.\\ \\When you output, use [START] before the output, and include [END] after the output. You may retain hash tags in the output only if hash tags were used in the original prompt. Here is the original description:}

\clearpage
\section{Dataset documentation}
\label{docu}
\noindent
\textbf{
This document is based on \textit{Datasheets for Datasets} by Gebru \textit{et
al.} \cite{gebru2021datasheets}.
}

\begin{mdframed}[linecolor=\sectioncolor]
\section*{\textcolor{\sectioncolor}{
    MOTIVATION
}}
\end{mdframed}

    \textcolor{\sectioncolor}{\textbf{
    For what purpose was the dataset created?
    }
    Was there a specific task in mind? Was there
    a specific gap that needed to be filled? Please provide a description.
    } \\
    This dataset consists of raw human-generated prompts that underwent human curation (via human voting to the showdown and pantheon channels), natural language captions generated using a large-language model on these prompts, multiple images generated using the natural langauge captions, and the semantic maps from diffusion cross-attention maps for the noun chunks parsed from the captions. To our knowledge, no other dataset has systematically recorded the cross-attention maps from a text-to-image generative model. Our dataset provides a basis for improving open vocabulary segmentation methods and for contrastive learning by leveraging the multiple images generated for each caption.
    
    \textcolor{\sectioncolor}{\textbf{
    Who created this dataset (e.g., which team, research group) and on behalf
    of which entity (e.g., company, institution, organization)?
    }
    } \\
    This dataset was created by a team from Carnegie Mellon University and leveraged CMU computational resources for the image generation. \\
    
    \textcolor{\sectioncolor}{\textbf{
    What support was needed to make this dataset?
    }
    (e.g.who funded the creation of the dataset? If there is an associated
    grant, provide the name of the grantor and the grant name and number, or if
    it was supported by a company or government agency, give those details.)
    } \\
    No specific grant was used. We utilized GPU computational resources available to CMU students for image generation. \\

\begin{mdframed}[linecolor=\sectioncolor]
\section*{\textcolor{\sectioncolor}{
    COMPOSITION
}}
\end{mdframed}
    \textcolor{\sectioncolor}{\textbf{
    What do the instances that comprise the dataset represent (e.g., documents,
    photos, people, countries)?
    }
    Are there multiple types of instances (e.g., movies, users, and ratings;
    people and interactions between them; nodes and edges)? Please provide a
    description.
    } \\
    The dataset is composed of the raw prompts generated by humans, the natural language captions derived from those prompts using a large language model, the ten images for each caption generated using a Stable Diffusion XL based model, and the individual semantic maps corresponding to the noun chunks for each caption for each generated image. \\
    
    \textcolor{\sectioncolor}{\textbf{
    How many instances are there in total (of each type, if appropriate)?
    }
    } \\
    There are 10 instances of generated images for each caption. Each of these captions has an average of ~5 noun chunks and thus 5 semantic maps for a generated image. Thus, on average, there are about 10 generated images and 50 semantic maps for each prompt. After NSFW and caption length filtering, 200 thousand such captions are used for image generation. Hence, about 2 million generated images and about 10 million semantic maps. We also share the raw versions of the prompts, thus there are also 224 thousand raw prompts (post NSFW filtering). \\
    
    \textcolor{\sectioncolor}{\textbf{
    Does the dataset contain all possible instances or is it a sample (not
    necessarily random) of instances from a larger set?
    }
    If the dataset is a sample, then what is the larger set? Is the sample
    representative of the larger set (e.g., geographic coverage)? If so, please
    describe how this representativeness was validated/verified. If it is not
    representative of the larger set, please describe why not (e.g., to cover a
    more diverse range of instances, because instances were withheld or
    unavailable).
    } \\
    We process the dataset to remove NSFW prompts and captions that exceed the Stable Diffusion XL 77 token limit from the image generation process.  \\
    
    \textcolor{\sectioncolor}{\textbf{
    What data does each instance consist of?
    }
    “Raw” data (e.g., unprocessed text or images) or features? In either case,
    please provide a description.
    } \\
    The generated image instances and the semantic maps are of "jpg" type. Both the raw prompts and natural language prompts are stored in a "pickle" file. \\
    
    \textcolor{\sectioncolor}{\textbf{
    Is there a label or target associated with each instance?
    }
    If so, please provide a description.
    } \\
    The generated image, corresponding semantic maps, the corresponding natural language text prompt, the raw prompt are associated with a unique hash key. This key helps identify and extract the corresponding instances. There are also "pickle" files present for each image that help identify the semantic map for a given noun chunk in the prompt. The seed used for image generation is also saved.\\
    
    \textcolor{\sectioncolor}{\textbf{
    Is any information missing from individual instances?
    }
    If so, please provide a description, explaining why this information is
    missing (e.g., because it was unavailable). This does not include
    intentionally removed information, but might include, e.g., redacted text.
    } \\
    We perform some simple heuristic-based filtering to remove certain noun chunks corresponding to non-physical objects. We only save the semantic maps for the remaining noun chunks. However, we will also provide the original captions so all noun chunks can be extracted if needed. \\
    
    \textcolor{\sectioncolor}{\textbf{
    Are relationships between individual instances made explicit (e.g., users’
    movie ratings, social network links)?
    }
    If so, please describe how these relationships are made explicit.
    } \\
    Yes, the relations are made explicit. The instances of generated images, prompts, and semantic maps are linked by a unique hash key. The relation between noun chunks and their corresponding semantic maps can be identified with the provided "offset" pickle file.  \\
    
    \textcolor{\sectioncolor}{\textbf{
    Are there recommended data splits (e.g., training, development/validation,
    testing)?
    }
    If so, please provide a description of these splits, explaining the
    rationale behind them.
    } \\
    No, there are no recommended splits. The user is free to split the data as per their need. \\
    
    \textcolor{\sectioncolor}{\textbf{
    Are there any errors, sources of noise, or redundancies in the dataset?
    }
    If so, please provide a description.
    } \\
    We observe that some raw prompts are highly similar, which seems to happen as a user quickly iterates on a prompt and creates prompts with a similar theme. We do not explicitly filter or remove such conceptually similar prompts. We also note that large language models are not perfect, and may create errors or deviations of the generated caption from the raw user prompts. \\
    
    \textcolor{\sectioncolor}{\textbf{
    Is the dataset self-contained, or does it link to or otherwise rely on
    external resources (e.g., websites, tweets, other datasets)?
    }
    If it links to or relies on external resources, a) are there guarantees
    that they will exist, and remain constant, over time; b) are there official
    archival versions of the complete dataset (i.e., including the external
    resources as they existed at the time the dataset was created); c) are
    there any restrictions (e.g., licenses, fees) associated with any of the
    external resources that might apply to a future user? Please provide
    descriptions of all external resources and any restrictions associated with
    them, as well as links or other access points, as appropriate.
    } \\
    It is self-contained and doesn't depend on any other resource. The resources used to create/modify the dataset have been mentioned above, however, they are no longer needed to use the dataset. \\
    
    \textcolor{\sectioncolor}{\textbf{
    Does the dataset contain data that might be considered confidential (e.g.,
    data that is protected by legal privilege or by doctor-patient
    confidentiality, data that includes the content of individuals’ non-public
    communications)?
    }
    If so, please provide a description.
    } \\
    No, the data was created by using publicly available information and doesn't involve any confidential or non-public information. \\
    
    \textcolor{\sectioncolor}{\textbf{
    Does the dataset contain data that, if viewed directly, might be offensive,
    insulting, threatening, or might otherwise cause anxiety?
    }
    If so, please describe why.
    } \\
    We have used Gemini 1.0 Pro to process the dataset to ensure that there are no harmful or offensive prompts. The prompts were then used to generate images from Stable Diffusion XL, which has an NSFW filter. The data is however subject to the failure of these models. \\
    
    \textcolor{\sectioncolor}{\textbf{
    Does the dataset relate to people?
    }
    If not, you may skip the remaining questions in this section.
    } \\
    As the prompts in the dataset were extracted from human written prompts in a discord server, the images generated and the prompts in the data may contain information about people.  \\
    
    \textcolor{\sectioncolor}{\textbf{
    Does the dataset identify any subpopulations (e.g., by age, gender)?
    }
    If so, please describe how these subpopulations are identified and
    provide a description of their respective distributions within the dataset.
    } \\
    The dataset is not targeted towards any subpopulation. However, it is subject to the distribution and preferences of the users of the discord server. \\
    
    \textcolor{\sectioncolor}{\textbf{
    Is it possible to identify individuals (i.e., one or more natural persons),
    either directly or indirectly (i.e., in combination with other data) from
    the dataset?
    }
    If so, please describe how.
    } \\
    The dataset is not targeted towards any individual, and we remove the information of the person submitting the prompt from the public release. The prompts themselves are subject to the distribution and preferences of the users of the discord server. \\
    
    \textcolor{\sectioncolor}{\textbf{
    Does the dataset contain data that might be considered sensitive in any way
    (e.g., data that reveals racial or ethnic origins, sexual orientations,
    religious beliefs, political opinions or union memberships, or locations;
    financial or health data; biometric or genetic data; forms of government
    identification, such as social security numbers; criminal history)?
    }
    If so, please provide a description.
    } \\
    The prompts used are extracted from users' input in the discord server. We have used Gemini 1.0 Pro to process the dataset to ensure that there are no harmful or offensive prompts. The prompts were then used to generate images from Stable Diffusion XL, which has an NSFW filter. The data is however subject to the failure of these models and the bias/preferences of the humans. \\

\begin{mdframed}[linecolor=\sectioncolor]
\section*{\textcolor{\sectioncolor}{
    COLLECTION
}}
\end{mdframed}

    \textcolor{\sectioncolor}{\textbf{
    How was the data associated with each instance acquired?
    }
    Was the data directly observable (e.g., raw text, movie ratings),
    reported by subjects (e.g., survey responses), or indirectly
    inferred/derived from other data (e.g., part-of-speech tags, model-based
    guesses for age or language)? If data was reported by subjects or
    indirectly inferred/derived from other data, was the data
    validated/verified? If so, please describe how.
    } \\
    The data was collected from the \texttt{showdown} and \texttt{pantheon} channels of the official Stability.ai discord server. In the discord server, users publicly posted prompts to obtain image generations and competed against other users for the chance to have prompts and images featured in the \texttt{pantheon} channel. We used \texttt{DiscordChatExporter} software to obtain the public prompts.\\
    
    \textcolor{\sectioncolor}{\textbf{
    Over what timeframe was the data collected?
    }
    Does this timeframe match the creation timeframe of the data associated
    with the instances (e.g., recent crawl of old news articles)? If not,
    please describe the timeframe in which the data associated with the
    instances was created. Finally, list when the dataset was first published.
    } \\
    This data was collected from the \texttt{showdown} channel from \textbf{July 11, 2023}, which is 1 day after the SDXL 1.0 candidates were made available via bots on discord; until \textbf{Feb 07, 2024} which is when the SDXL bots were shut down. As \texttt{pantheon} was derived from \texttt{showdown}, and maintained the history from the start of channel creation, we scraped the \texttt{pantheon} channel only once, which collected posts from \textbf{May 02, 2023} onwards.\\
    
    \textcolor{\sectioncolor}{\textbf{
    What mechanisms or procedures were used to collect the data (e.g., hardware
    apparatus or sensor, manual human curation, software program, software
    API)?
    }
    How were these mechanisms or procedures validated?
    } \\
    We used \texttt{DiscordChatExporter} executed on a Linux machine on a $14$ minute cron job to capture all posts from \texttt{showdown}. The \texttt{showdown} was updated every $15$ minutes, with history cleared every $30$ minutes.
    
    \textcolor{\sectioncolor}{\textbf{
    What was the resource cost of collecting the data?
    }
    (e.g. what were the required computational resources, and the associated
    financial costs, and energy consumption - estimate the carbon footprint.)
    } \\
    We executed the data collection from Discord on a machine that was on regardless of usage, we do not believe this imposed significant CPU usage or energy cost. \\
    
    \textcolor{\sectioncolor}{\textbf{
    If the dataset is a sample from a larger set, what was the sampling
    strategy (e.g., deterministic, probabilistic with specific sampling
    probabilities)?
    }
    } \\
    We filtered the dataset using a large language model, which caused around $4.6\%$ of the 235k prompts to be removed due to safety reasons (not transformed to natural language captions), yielding 224k natural language captions. Furthermore, captions longer than $77$ tokens were not used for image generation, and roughly 200k captions were used for images. \\
    
    \textcolor{\sectioncolor}{\textbf{
    Who was involved in the data collection process (e.g., students,
    crowdworkers, contractors) and how were they compensated (e.g., how much
    were crowdworkers paid)?
    }
    } \\
    The data collection was performed by students as part of their research. \\
    
    \textcolor{\sectioncolor}{\textbf{
    Were any ethical review processes conducted (e.g., by an institutional
    review board)?
    }
    If so, please provide a description of these review processes, including
    the outcomes, as well as a link or other access point to any supporting
    documentation.
    } \\
    No IRB was required as this data collection did not directly involve human subjects. \\
    
    \textcolor{\sectioncolor}{\textbf{
    Does the dataset relate to people?
    }
    If not, you may skip the remainder of the questions in this section.
    } \\
    Not directly, however, the prompts were submitted by users. \\
    
    \textcolor{\sectioncolor}{\textbf{
    Did you collect the data from the individuals in question directly, or
    obtain it via third parties or other sources (e.g., websites)?
    }
    } \\
    They were collected from the chat messages publicly submitted by users to Discord. \\
    
    \textcolor{\sectioncolor}{\textbf{
    Were the individuals in question notified about the data collection?
    }
    If so, please describe (or show with screenshots or other information) how
    notice was provided, and provide a link or other access point to, or
    otherwise reproduce, the exact language of the notification itself.
    } \\
    No notification was provided as the messages were publicly shared, and expected to be visible and voted upon by other users. We notified moderators of the Stability Discord server that we were performing this data collection effort and did not receive any objection. \\
    
    \textcolor{\sectioncolor}{\textbf{
    Did the individuals in question consent to the collection and use of their
    data?
    }
    If so, please describe (or show with screenshots or other information) how
    consent was requested and provided, and provide a link or other access
    point to, or otherwise reproduce, the exact language to which the
    individuals consented.
    } \\
    Users consented to their prompts being publicly visible by submitting prompts for image generation. Other users were explicitly encouraged to view and vote on the prompts and images. \\
    
    \textcolor{\sectioncolor}{\textbf{
    Has an analysis of the potential impact of the dataset and its use on data
    subjects (e.g., a data protection impact analysis)been conducted?
    }
    If so, please provide a description of this analysis, including the
    outcomes, as well as a link or other access point to any supporting
    documentation.
    } \\
    We performed an NSFW evaluation using Llama Guard 2 and found that our captions post-filtering were overwhelmingly safe. Visual examination of over a thousand randomly selected captions also confirmed that our captions were largely safe.\\

\begin{mdframed}[linecolor=\sectioncolor]
\section*{\textcolor{\sectioncolor}{
    PREPROCESSING / CLEANING / LABELING
}}
\end{mdframed}

    \textcolor{\sectioncolor}{\textbf{
    Was any preprocessing/cleaning/labeling of the data
    done(e.g.,discretization or bucketing, tokenization, part-of-speech
    tagging, SIFT feature extraction, removal of instances, processing of
    missing values)?
    }
    If so, please provide a description. If not, you may skip the remainder of
    the questions in this section.
    } \\
    We performed our prompt cleaning in three steps. \textbf{(1)} We use a set of handcrafted \texttt{grep} filters to remove common prompt corruptions, mispunctuation, extra spaces, and non-text Unicode. \textbf{2} We deduplicate the prompts after \texttt{grep} filtering. \textbf{(3)} These prompts are provided to a large language model, where they are filtered and transformed into natural language captions.\\

    \textcolor{\sectioncolor}{\textbf{
    Was the “raw” data saved in addition to the preprocessed/cleaned/labeled
    data (e.g., to support unanticipated future uses)?
    }
    If so, please provide a link or other access point to the “raw” data.
    } \\
    Yes, the raw prompts will be shared. \\

    \textcolor{\sectioncolor}{\textbf{
    Is the software used to preprocess/clean/label the instances available?
    }
    If so, please provide a link or other access point.
    } \\
    Yes, the code will be publicly provided after paper acceptance. \\

\begin{mdframed}[linecolor=\sectioncolor]
\section*{\textcolor{\sectioncolor}{
    USES
}}
\end{mdframed}

    \textcolor{\sectioncolor}{\textbf{
    Has the dataset been used for any tasks already?
    }
    If so, please provide a description.
    } \\
    We use our dataset to evaluate captioning and open vocabulary segmentation models. However, the dataset has not been used to train new models as of writing. \\

    \textcolor{\sectioncolor}{\textbf{
    Is there a repository that links to any or all papers or systems that use the dataset?
    }
    If so, please provide a link or other access point.
    } \\
    The link to the dataset will contain papers that use our dataset: \url{https://stablesemantics.github.io/StableSemantics/}  \\

    \textcolor{\sectioncolor}{\textbf{
    What (other) tasks could the dataset be used for?
    }
    } \\
    The dataset can be used for training/finetuning and evaluating scene understanding, semantic segmentation, and object detection/classification models. It can also be used for developing models for tasks like Visual Question Answering (VQA), visual grounding, inpainting, etc. It can also be used for understanding the distribution of natural language prompts and the biases present in them.  \\

    \textcolor{\sectioncolor}{\textbf{
    Is there anything about the composition of the dataset or the way it was
    collected and preprocessed/cleaned/labeled that might impact future uses?
    }
    For example, is there anything that a future user might need to know to
    avoid uses that could result in unfair treatment of individuals or groups
    (e.g., stereotyping, quality of service issues) or other undesirable harms
    (e.g., financial harms, legal risks) If so, please provide a description.
    Is there anything a future user could do to mitigate these undesirable
    harms?
    } \\
    The raw prompts collected are human written prompts and are then processed by Gemini 1.0 Pro. These prompts are then used to generate the images from Stable Diffusion XL based model. These models were used following their license terms. \\

    \textcolor{\sectioncolor}{\textbf{
    Are there tasks for which the dataset should not be used?
    }
    If so, please provide a description.
    } \\
    Preventative steps have been taken to ensure there is no harmful content in the data as explained in the data processing part. However, we still recommend the data should not be used for any harmful or illegal purpose. \\

\begin{mdframed}[linecolor=\sectioncolor]
\section*{\textcolor{\sectioncolor}{
    DISTRIBUTION
}}
\end{mdframed}

    \textcolor{\sectioncolor}{\textbf{
    Will the dataset be distributed to third parties outside of the entity
    (e.g., company, institution, organization) on behalf of which the dataset
    was created?
    }
    If so, please provide a description.
    } \\
    The dataset will be under CC0 1.0 Universal Public Domain Dedication license and thus will distributed openly without any restriction. \\

    \textcolor{\sectioncolor}{\textbf{
    How will the dataset be distributed (e.g., tarball on website, API,
    GitHub)?
    }
    Does the dataset have a digital object identifier (DOI)?
    } \\
    The dataset will be shared via a torrent and hosted via http(s). The links to access the dataset will be available on our website which will be always maintained. \\

    \textcolor{\sectioncolor}{\textbf{
    When will the dataset be distributed?
    }
    } \\
    The complete dataset will be made publicly available by the end of June 2024. \\

    \textcolor{\sectioncolor}{\textbf{
    Will the dataset be distributed under a copyright or other intellectual
    property (IP) license, and/or under applicable terms of use (ToU)?
    }
    If so, please describe this license and/or ToU, and provide a link or other
    access point to, or otherwise reproduce, any relevant licensing terms or
    ToU, as well as any fees associated with these restrictions.
    } \\
    It will not be under any copyright or intellectual property (IP) license. The dataset will be released under the CC0 1.0 Universal Public Domain Dedication license.  \\

    \textcolor{\sectioncolor}{\textbf{
    Have any third parties imposed IP-based or other restrictions on the data
    associated with the instances?
    }
    If so, please describe these restrictions, and provide a link or other
    access point to, or otherwise reproduce, any relevant licensing terms, as
    well as any fees associated with these restrictions.
    } \\
    No. \\

    \textcolor{\sectioncolor}{\textbf{
    Do any export controls or other regulatory restrictions apply to the
    dataset or to individual instances?
    }
    If so, please describe these restrictions, and provide a link or other
    access point to, or otherwise reproduce, any supporting documentation.
    } \\
    No. \\

\begin{mdframed}[linecolor=\sectioncolor]
\section*{\textcolor{\sectioncolor}{
    MAINTENANCE
}}
\end{mdframed}

    \textcolor{\sectioncolor}{\textbf{
    Who is supporting/hosting/maintaining the dataset?
    }
    } \\
    The authors will be hosting and maintaining the dataset. \\

    \textcolor{\sectioncolor}{\textbf{
    How can the owner/curator/manager of the dataset be contacted (e.g., email
    address)?
    }
    } \\
    The authors should be contacted via email for any questions/problems regarding the dataset. \\

    \textcolor{\sectioncolor}{\textbf{
    Is there an erratum?
    }
    If so, please provide a link or other access point.
    } \\
    There is no identified erratum with the dataset so far. However, the user must consider any possible errors associated with the LLM and Diffusion models used. \\

    \textcolor{\sectioncolor}{\textbf{
    Will the dataset be updated (e.g., to correct labeling errors, add new
    instances, delete instances)?
    }
    If so, please describe how often, by whom, and how updates will be
    communicated to users (e.g., mailing list, GitHub)?
    } \\
    The dataset will be updated by the authors if there is any error detected. The information will be announced on our website from where people can access the dataset.\\

    \textcolor{\sectioncolor}{\textbf{
    If the dataset relates to people, are there applicable limits on the
    retention of the data associated with the instances (e.g., were individuals
    in question told that their data would be retained for a fixed period of
    time and then deleted)?
    }
    If so, please describe these limits and explain how they will be enforced.
    } \\
    The dataset doesn't collect any individual's personal information. \\

    \textcolor{\sectioncolor}{\textbf{
    Will older versions of the dataset continue to be
    supported/hosted/maintained?
    }
    If so, please describe how. If not, please describe how its obsolescence
    will be communicated to users.
    } \\
    The older versions of the dataset will continue to be supported. Any updates made in the dataset will be hosted as separate instances so that users of the previous version can still access it. \\

    \textcolor{\sectioncolor}{\textbf{
    If others want to extend/augment/build on/contribute to the dataset, is
    there a mechanism for them to do so?
    }
    If so, please provide a description. Will these contributions be
    validated/verified? If so, please describe how. If not, why not? Is there a
    process for communicating/distributing these contributions to other users?
    If so, please provide a description.
    } \\
    We have no restriction on others augmenting this dataset. However, we don't aim to officially merge the extended/augmented data to avoid any possible inclusion of offensive/harmful or illegal instances. \\

\clearpage
\section{Broader Impacts}
\label{impact}
StableSemantics consists of raw prompts, natural language captions derived from the prompts, $10$ images per caption generated using a Stable Diffusion XL based model, and semantic maps corresponding to the noun chunks in the caption for each image generation. The availability of this dataset is expected to improve visual semantic learning, open vocabulary segmentation methods, and improve the characterization of text-to-image generative models. We expect StableSemantics to help develop models that can understand complex visual scenes. This has the potential to increase the robustness of deep learning-based visual systems. 
\section{Dataset website}
\label{website}
Note that the dataset will be soon released. For now, we have released a sample consisting of images corresponding to $100$ samples of $5$ images each on the website. \url{https://stablesemantics.github.io/StableSemantics/}
\section{Statement of responsibility}
\label{responsibility}\unskip
Authors accept all responsibility for potential violation of rights.
\section{Dataset license}
\label{license}\unskip
We release our dataset under a \texttt{CC0 1.0 Universal Public Domain Dedication license} following Stable Diffusion's \href{https://stability.ai/discord-tos}{discord policy} and other work collecting data from Stable Diffusion's discord~\cite{wangDiffusionDBLargescalePrompt2022}.
\section{Hosting plan and dataset format}
~\label{hosting}\unskip
We plan to host our dataset via an http(s) link, and simultaneously distribute the data via a torrent where we permanently seed the files. The captions are stored in a Python pickle file, while images and semantic maps are jpg files.
\clearpage


\end{document}